\documentclass[letterpaper, 10 pt, conference]{ieeeconf}  

\IEEEoverridecommandlockouts                              
\makeatletter
\let\NAT@parse\undefined
\makeatother

\usepackage[dvipsnames]{xcolor}

\newcommand*\linkcolours{ForestGreen}

\usepackage{times}
\usepackage{graphicx}
\usepackage{amssymb}
\usepackage{gensymb}
\usepackage{tikz}
\usetikzlibrary{positioning, shapes.geometric, arrows.meta}

\usepackage{amsmath}
\usepackage{breakurl}
\usepackage{caption}
\usepackage{subcaption}
\usepackage{cuted}

\usepackage[linesnumbered,ruled,vlined]{algorithm2e}

\usepackage[utf8]{inputenc}

\usepackage[figuresright]{rotating}
\usepackage{booktabs} 
\usepackage{multirow} 
\usepackage{tikz}
\usetikzlibrary{positioning, arrows.meta, shapes, calc}

\usepackage{url,hyperref}
\hypersetup{
colorlinks,
linkcolor=\linkcolours,
citecolor=\linkcolours,
filecolor=\linkcolours,
urlcolor=\linkcolours}

\usepackage[labelfont={bf},font=small]{caption}
\usepackage[none]{hyphenat}
\usepackage{tikz}
\usetikzlibrary{positioning, shapes.geometric, arrows.meta, fit, backgrounds}
\usetikzlibrary{positioning, shapes.geometric, arrows.meta, fit, backgrounds}
\usepackage{mathtools, cuted}

\usepackage[noadjust, nobreak]{cite}
\usepackage{tabularx}
\usepackage{amsmath}
\usepackage{float}
\usepackage{pifont}

\newcolumntype{Y}{>{\centering\arraybackslash}X}
\usepackage[]{placeins}

\usepackage{placeins}

\usepackage{tikz}

\usepackage[framemethod=tikz]{mdframed}
\usepackage{afterpage}
\usepackage{stfloats}
\usepackage{atbegshi}
\usepackage{amsmath}
\usepackage{graphicx}
\usepackage{subcaption}
\usepackage{multirow}
\newcommand{\handlethispage}{}
\newcommand{\discardpagesfromhere}{\let\handlethispage\AtBeginShipoutDiscard}
\newcommand{\keeppagesfromhere}{\let\handlethispage\relax}
\AtBeginShipout{\handlethispage}

\usepackage{comment}
\title{Adaptive Head Budgeting for Efficient Multi-Head Attention}

\author{Bilal FAYE$^{1}$, Abdoulaye MBAYE$^{2}$, Hanane AZZAG$^{3}$, Mustapha Lebbah$^{4}$  \\
	\normalsize e-mail: faye@lipn.univ-paris13.fr, a.mbaye@yobbal.com, azzag@univ-paris13.fr, mustapha.lebbah@uvsq.fr
}

\begin{document}

\maketitle
\thispagestyle{empty}
\pagestyle{empty}

\begin{abstract}
Multi-head attention enables Transformers to capture diverse representations, but all attention heads are typically activated for every input, regardless of task complexity. For coarse-grained tasks such as text classification, where relevant information is often global, this fixed allocation can introduce unnecessary computation. We propose \textbf{BudgetFormer}, a Transformer architecture that dynamically allocates attention heads on a per-input basis. The model learns both a head budget and a relevance distribution to select the most informative heads. To support effective head selection, we introduce a training strategy that balances exploration and exploitation. Experiments on text classification tasks show that BudgetFormer reduces FLOPs and memory usage while matching or surpassing the performance of standard multi-head attention. These results highlight adaptive head allocation as an effective approach to improving Transformer efficiency and performance.
\end{abstract}

\section{Introduction}
Transformers have become the dominant architecture in natural language processing and beyond, driven by the effectiveness of self-attention mechanisms in modeling long-range dependencies \cite{vaswani2017attention,devlin2019bert}. In particular, multi-head attention enables the model to capture diverse representation subspaces by projecting inputs into multiple parallel attention heads. This design has been central to the success of large-scale models across tasks such as language understanding, generation, and classification.\newline

\noindent However, the computational cost of self-attention scales quadratically with the sequence length, making it a major bottleneck in practice~\cite{vaswani2017attention}. This limitation becomes especially pronounced in autoregressive generation, where tokens are processed sequentially and inference latency accumulates over time. To mitigate this issue, techniques such as key-value caching are commonly used to reuse past computations and reduce redundant operations during decoding~\cite{dai2019transformer,child2019generating}. Despite these optimizations, the cost of attention remains significant, particularly in large models and long-context settings.\newline

\noindent A broad range of methods has been proposed to improve the efficiency of Transformers. Model compression techniques such as knowledge distillation reduce model size while preserving performance~\cite{sanh2019distilbert}. Sparse and approximate attention mechanisms aim to alleviate the quadratic complexity by restricting attention patterns or using low-rank approximations~\cite{beltagy2020longformer,zaheer2020big,choromanski2021performer}. Other approaches include token pruning, which removes less informative tokens during inference, and early exiting strategies that adaptively reduce the depth of computation~\cite{hou2020dynabert,liu2021faster}. While effective in certain settings, these methods often require architectural modifications, introduce approximation errors, or rely on heuristics that may not generalize well across tasks.\newline

\noindent In this work, we focus on a complementary and largely underexplored dimension of efficiency: the adaptive use of attention heads. Standard multi-head attention activates all heads uniformly for every input, regardless of its complexity or the nature of the task. This can be suboptimal, especially in coarse-grained tasks such as text classification, where the relevant information is often global and does not require the full diversity of attention heads. Using a fixed number of heads may therefore lead to unnecessary computation or inefficient allocation of model capacity.\newline

\noindent To address this limitation, we introduce BudgetFormer, a Transformer architecture equipped with an adaptive multi-head attention mechanism that dynamically allocates computational resources at the head level. For each input, the model estimates a head budget corresponding to the number of attention heads required, and selects the most informative heads based on learned relevance scores. In addition, we propose a training strategy based on an exploration and exploitation trade-off, allowing the model to effectively discover and refine head usage patterns. Our contributions can be summarized as follows:
\begin{itemize}
    \item We propose an adaptive multi-head attention mechanism that learns to allocate a variable number of attention heads per input based on its complexity.
    \item We introduce a training strategy that balances exploration and exploitation to learn efficient and robust head selection policies.
    \item We demonstrate that our approach reduces inference cost in terms of FLOPs and memory usage, leading to more frugal models with lower computational and environmental footprint.
    \item We validate our method on text classification tasks of varying complexity, showing that BudgetFormer can outperform standard full multi-head attention while using fewer computational resources.
\end{itemize}

\section{Related Work}

Improving the efficiency of Transformer models has become a major research direction due to the high computational and memory cost of self-attention. Existing approaches can be broadly categorized into model compression, sparse and approximate attention, token-level adaptivity, and conditional computation mechanisms. While these methods have shown promising results, they often operate at the level of tokens, layers, or full attention maps, leaving the adaptive allocation of attention heads relatively underexplored.

\subsection{Model Compression and Pruning}
Model compression techniques aim to reduce the size and computational cost of Transformers while preserving performance. Knowledge distillation methods train smaller student models to mimic larger teachers~\cite{sanh2019distilbert}. Structured pruning approaches remove redundant components such as weights, neurons, or attention heads based on importance criteria~\cite{ganesh2020compressing}.\newline

\noindent Head pruning in particular has revealed that many attention heads are redundant and can be removed with minimal performance degradation. For instance, pruning strategies based on search or saliency metrics can eliminate a significant fraction of heads without loss in accuracy~\cite{parnami2021pruning}. More recent works extend this idea by combining head pruning with block or token pruning, highlighting the redundancy present in both attention maps and head structures~\cite{jaradat2024hybrid}.\newline

\noindent However, these approaches are typically static, requiring pruning decisions to be made offline and applied uniformly across all inputs. This limits their ability to adapt computation dynamically based on input complexity.

\subsection{Sparse and Approximate Attention}
Another line of work focuses on reducing the quadratic complexity of self-attention by introducing sparsity or approximation. Methods such as Longformer, BigBird, and Performer replace dense attention with structured or kernel-based approximations~\cite{beltagy2020longformer,zaheer2020big,choromanski2021performer}. These approaches achieve sub-quadratic complexity while maintaining strong empirical performance.\newline

\noindent Despite their efficiency gains, sparse attention methods often rely on predefined patterns or approximations that may restrict the expressiveness of the model. In particular, fixed sparsity structures can limit the ability to capture global dependencies when needed~\cite{beltagy2020longformer}. 

\subsection{Token Pruning and Adaptive Sequence Processing}

Token-level methods aim to reduce computation by dynamically selecting or pruning tokens during inference. Techniques such as PoWER-BERT and subsequent works remove less informative tokens based on learned importance scores. More recent approaches introduce progressive or dynamic token pruning strategies that adaptively refine the token set across layers~\cite{goyal2020power}.\newline

\noindent Recent work has further explored adaptive token retention and pruning in both NLP and vision settings, demonstrating significant reductions in FLOPs while maintaining accuracy~\cite{liao2024catp}. Additionally, dynamic pruning methods have been proposed to jointly prune tokens, heads, and attention blocks during inference, highlighting the redundancy present across multiple dimensions of the Transformer architecture~\cite{jaradat2024hybrid}.\newline

\noindent However, token pruning methods may struggle in tasks where all tokens contribute to the final prediction, such as fine-grained reasoning or dense prediction tasks. Moreover, pruning decisions are often irreversible within a forward pass, which can lead to information loss.

\subsection{Conditional Computation and Early Exiting}
Conditional computation approaches aim to adapt the amount of computation to the difficulty of each input. Early exiting methods allow models to produce predictions at intermediate layers, reducing average inference depth~\cite{zhou2020bert}. Similarly, adaptive-depth Transformers dynamically select the number of layers to apply per input, achieving substantial reductions in computation \cite{liu2021faster}.\newline

\noindent Mixture-of-Experts models extend this idea by routing inputs to a subset of expert networks, enabling scalable conditional computation~\cite{fedus2022switch}. More recent works also explore skipping layers or dynamically adjusting network depth based on input complexity~\cite{lawson2025learning}.\newline

\noindent While effective, these approaches primarily operate at the level of layers or feed-forward modules. They do not explicitly address the allocation of attention heads within each layer, which remains fixed in standard architectures.

\subsection{Discussion}

Across these lines of work, a common theme is the presence of significant redundancy in Transformer computations, whether at the level of tokens, layers, or attention structures. In particular, recent analyses show that only a subset of attention heads contributes meaningfully to global information processing, while many heads focus on local or redundant patterns.\newline

\noindent Despite this observation, existing methods either remove heads statically or treat all heads uniformly during inference. This suggests a gap in current approaches: the lack of fine-grained, input-dependent allocation of attention heads.\newline
In contrast, our approach focuses on dynamically allocating attention heads on a per-input basis. Rather than pruning or approximating attention globally, we learn to estimate a head budget and select the most relevant heads for each input. This enables a more flexible and fine-grained form of conditional computation that is particularly well suited for coarse-grained tasks such as classification, where the required level of attention diversity may vary significantly across inputs.

\section{Background}

\subsection{Multi-Head Self-Attention}

Let $X \in \mathbb{R}^{B \times N \times D}$ denote a sequence of $N$ input tokens, where $B$ is the batch size and $D$ the model dimension. In Transformer encoders, self-attention operates by projecting $X$ into queries, keys, and values through linear mappings:
\begin{equation}
Q = X W_Q,\quad K = X W_K,\quad V = X W_V
\end{equation}
with $W_Q, W_K, W_V \in \mathbb{R}^{D \times D}$. These representations are partitioned into $H$ heads, each of dimension $d_h = D / H$, allowing the model to attend to information from multiple representation subspaces.\newline

\noindent For each head $h$, attention is computed as:
\begin{equation}
\text{Attn}_h(X) = \text{Softmax}\left( \frac{Q_h K_h^\top}{\sqrt{d_h}} \right) V_h
\end{equation}
The outputs of all heads are concatenated and projected back to the model dimension:
\begin{equation}
\text{MHA}(X) = \text{Concat}(\text{Attn}_1(X), \dots, \text{Attn}_H(X)) W_O
\end{equation}
where $W_O$ is the output projection matrix.\newline

\noindent This mechanism allows each head to capture distinct interaction patterns across the sequence, which has been identified as a key factor behind the empirical success of Transformer models \cite{vaswani2017attention}.

\subsection{Computational Complexity}

The computational cost of multi-head attention arises from both projection operations and pairwise interactions between tokens. Given an input of length $N$, the computation of attention scores involves forming the matrix product $Q_h K_h^\top \in \mathbb{R}^{N \times N}$ for each head. This induces a quadratic dependency on the sequence length.\newline

\noindent More precisely, for a single layer, the dominant cost can be expressed as:
\begin{equation}
\mathcal{C}_{\text{attn}}(X) \approx \mathcal{O}(B N D^2) + \mathcal{O}(B H N^2 d_h)
\end{equation}
where the first term corresponds to the linear projections and the second to the attention computation and aggregation. Using $D = H d_h$, the quadratic term becomes $\mathcal{O}(B N^2 D)$, which dominates for long sequences.\newline

\noindent An important observation is that this cost scales linearly with the number of heads $H$. All heads are computed independently, and their contributions are aggregated uniformly, regardless of their individual relevance to the input. As a result, increasing $H$ improves representational capacity but also directly increases computational cost.

\subsection{Memory Requirements}

Beyond computation, memory consumption is another critical factor in Transformer models. In encoder architectures, all tokens are processed simultaneously, and intermediate attention representations must be stored during the forward pass.\newline

\noindent The attention score tensors for each head have size $\mathbb{R}^{N \times N}$, leading to a total storage cost proportional to $B H N^2$. In addition, attention probabilities, intermediate projections, and output representations contribute linearly in $B N D$.\newline

\noindent During training, these activations must be retained for backpropagation, effectively doubling memory usage. Consequently, the overall memory footprint of a Transformer encoder layer is dominated by the quadratic term:
\begin{equation}
\mathcal{M}_{\text{attn}}(X) \approx \mathcal{O}(B H N^2)
\end{equation}
This scaling makes attention particularly expensive in settings with long sequences or large numbers of heads.

\subsection{Inference Efficiency in Encoder Models}

In encoder-based tasks such as text classification, inference typically processes the entire sequence in a single forward pass. While this avoids the sequential overhead of autoregressive decoding, the full attention computation remains necessary for all tokens and all heads.\newline

\noindent In this setting, the computational cost per layer remains proportional to $H N^2$, and the model evaluates all attention heads regardless of the input structure. However, not all inputs require the same level of representational diversity. For instance, in coarse-grained classification tasks, the decision often relies on global semantic cues that can be captured by a subset of attention heads.\newline

\noindent This suggests that the uniform use of all heads may lead to over-computation, where some heads contribute marginally to the final representation while still incurring full computational and memory cost.

\subsection{Motivation}

The above analysis highlights two structural inefficiencies in standard multi-head attention. First, the cost of attention grows linearly with the number of heads, making head multiplicity a direct driver of computational and memory overhead. Second, the architecture assumes that all heads are equally useful for every input, which is unlikely to hold in practice, especially in tasks where the required level of abstraction varies across examples.\newline

\noindent These observations motivate the design of adaptive mechanisms that can modulate the number of active heads depending on the input. Instead of treating all heads uniformly, it becomes natural to consider a formulation in which only a subset of heads is selected or weighted more strongly, reducing unnecessary computation while preserving task-relevant information.\newline

\noindent In the next section, we build on this perspective and introduce an adaptive attention mechanism that learns to allocate a head budget and select informative heads dynamically.

\section{Method: BudgetFormer}
\label{section:method}
In this section, we introduce BudgetFormer, a Transformer encoder equipped with adaptive head-level computation. Unlike standard multi-head attention, which activates all heads uniformly for every input, our approach learns to dynamically allocate a computational budget over attention heads. This allows the model to adapt its level of computation to the complexity of each input, while maintaining full model capacity during training.

\subsection{Adaptive Head Budgeted Attention}

Let $X \in \mathbb{R}^{B \times N \times D}$ denote the input representation, where $B$ is the batch size, $N$ the sequence length, and $D$ the model dimension. A global summary is obtained for each sample by mean pooling over the token dimension:
\begin{equation}
h_b = \frac{1}{N}\sum_{n=1}^{N} X_{b,n},
\qquad b=1,\ldots,B,
\end{equation}
where $h_b \in \mathbb{R}^{D}$.\newline

\noindent The budget network predicts an input-dependent control variable:
\begin{equation}
s_b = \sigma\!\left(f_{\theta}(h_b)\right),
\end{equation}
where $f_{\theta}: \mathbb{R}^{D} \rightarrow \mathbb{R}$ and $s_b \in [0,1]$ represents the fraction of attention heads to activate.\newline

\noindent Head relevance scores are then computed as
\begin{equation}
z_b = g_{\phi}(h_b)
      + \epsilon_b \cdot \sigma_{\max}\left(1-\frac{t}{T}\right),
\end{equation}
where $g_{\phi}: \mathbb{R}^{D} \rightarrow \mathbb{R}^{H}$, $\epsilon_b \sim \mathcal{N}(0, I_H)$, $t$ and $T$ denote the current and total training steps, respectively, $\sigma_{\max}$ controls the initial exploration noise scale, and $H$ is the number of attention heads.\newline

\noindent A probability distribution over heads is obtained through a temperature-scaled softmax:
\begin{equation}
p_{b,h}
=
\frac{\exp(z_{b,h}/\tau(t))}
{\sum_{j=1}^{H}\exp(z_{b,j}/\tau(t))},
\end{equation}
with
\begin{equation}
\tau(t)
=
\tau_{\min}
+
(\tau_{\max}-\tau_{\min})
\exp\!\left(-\gamma \frac{t}{T}\right).
\end{equation}
The importance assigned to head $h$ for sample $b$ is
\begin{equation}
w_{b,h}
=
s_b\, H\, p_{b,h}.
\end{equation}
For each attention head,
\begin{equation}
A_{b,h}
=
\mathrm{Softmax}
\left(
\frac{Q_{b,h}K_{b,h}^{\top}}
{\sqrt{d_h}}
\right)
V_{b,h},
\end{equation}
where $d_h=D/H$.
The weighted head outputs are
\begin{equation}
\tilde{A}_{b,h}
=
w_{b,h} A_{b,h}.
\end{equation}
The final attention output is
\begin{equation}
Y_b
=
\mathrm{Concat}
\left(
\tilde{A}_{b,1},
\dots,
\tilde{A}_{b,H}
\right)
W_O.
\end{equation}
During inference, only the top-$k_b$ heads are retained:
\begin{equation}
k_b
=
\max
\left(
1,
\left\lfloor s_b H \right\rfloor
\right).
\end{equation}
Let $\mathcal{S}_{b}^{(k_b)}$ denote the set of indices corresponding to the $k_b$ largest values of $p_{b,h}$. A binary mask is defined as
\begin{equation}
m_{b,h}
=
\begin{cases}
1, & h \in \mathcal{S}_{b}^{(k_b)},\\
0, & \text{otherwise}.
\end{cases}
\end{equation}
Only active heads are evaluated:
\begin{equation}
\tilde{A}_{b,h}
=
\begin{cases}
w_{b,h}A_{b,h}, & m_{b,h}=1,\\
0, & m_{b,h}=0.
\end{cases}
\end{equation}
The resulting output is
\begin{equation}
Y_b
=
\mathrm{Concat}
\left(
\tilde{A}_{b,1},
\dots,
\tilde{A}_{b,H}
\right)
W_O.
\end{equation}
Thus, the number of evaluated heads is reduced from $H$ to $k_b$ for each input sample $X_b$.\newline

\noindent The combination of the noise scale $\sigma_{\max}$ and the temperature schedule $\tau(t)$ defines a gradual transition from exploration to exploitation, while the budget variable $s$ controls the global computational allocation per input.

\subsection{Training Objective}
Optimizing only the task loss is insufficient in our setting, as it does not constrain how computational resources are allocated across attention heads. In particular, the model may converge to degenerate solutions where all heads are uniformly used or where the budget collapses to extreme values.\newline

\noindent We therefore define the following objective:
\begin{equation}
\mathcal{L} = \mathcal{L}_{task} + \mathcal{L}_{budget} + \mathcal{L}_{entropy},
\end{equation}
where $\mathcal{L}_{task}$ is task-dependent (e.g., classification or regression), $\mathcal{L}_{budget}$ controls the global allocation of heads, and $\mathcal{L}_{entropy}$ regulates head specialization.\newline

\noindent The budget $s_b$ is constrained within a target interval $[s_{\min}, s_{\max}]$ using a quadratic hinge formulation. We first define the violation as:
\begin{equation}
v(s_b) = \max(0, s_{\min} - s_b) + \max(0, s_b - s_{\max}).
\end{equation}
The budget loss is then given by:
\begin{equation}
\mathcal{L}_{budget} = \alpha(s_b) \cdot v(s_b)^2,
\end{equation}
where the scaling factor adapts to the magnitude of the violation:
\begin{equation}
\alpha(s_b) = \min(\alpha_{\max}, \alpha_{\text{base}} + v(s_b)).
\end{equation}
This formulation allows the model to freely explore any value of $s_b$ within the interval without penalty, while progressively increasing the constraint when $s_b$ deviates from the desired range. The adaptive scaling prevents unstable behavior and avoids collapse toward trivial budgets.\newline

\noindent To control the distribution over heads, we introduce an entropy regularization term:
\begin{equation}
\mathcal{L}_{entropy} = \sum_{h=1}^{H} p_{b,h} \log p_{b,h},
\end{equation}
where $p_{b,h}$ are the head selection probabilities. Its influence is modulated over training through a time-dependent coefficient:
\begin{equation}
\beta(t) = \beta_{\max} \left( \frac{2t}{T} - 1 \right).
\end{equation}
At early stages ($t \approx 0$), $\beta(t) < 0$, which favors high-entropy distributions and encourages exploration across heads. Around mid-training, $\beta(t) \approx 0$, reducing its effect. At later stages ($t > T/2$), $\beta(t) > 0$, which promotes low-entropy distributions and leads to sparse and specialized head usage.
The entropy term is thus defined as:
\begin{equation}
\mathcal{L}_{entropy} = \beta(t) \sum_{h=1}^{H} p_h \log p_h.
\end{equation}
The combination of the violation-based budget constraint and the entropy schedule enables a stable transition from exploration to exploitation, while explicitly controlling the computational footprint of the model.\newline

\begin{algorithm}
\caption{BudgetFormer Training Procedure}
\label{alg:budgetformer}

\KwIn{Dataset $\mathcal{D} = \{(X_i, y_i)\}_{i=1}^N$, parameters $\theta, \phi$, attention parameters $W_Q, W_K, W_V, W_O$, number of heads $H$, total steps $T$, learning rate $\eta$}
\KwOut{Trained parameters $\theta, \phi, W_Q, W_K, W_V, W_O$}

\For{$t \leftarrow 1$ \KwTo $T$}{
    \ForEach{mini-batch $B \subset \mathcal{D}$}{

        \ForEach{$(X, y) \in B$}{

            Compute token embeddings $X \in \mathbb{R}^{B \times N \times D}$\;

            \tcp{Global summary}
            Compute $h_b \leftarrow \frac{1}{N}\sum_{n=1}^{N} X_{b,n}$\;

            \tcp{Budget prediction}
            Compute $s_b \leftarrow \sigma(f_{\theta}(h_b))$\;

            \tcp{Head scoring with exploration noise}
            Sample $\epsilon_b \sim \mathcal{N}(0, I_H)$\;
            Compute $z_b \leftarrow g_{\phi}(h_b) + \epsilon_b \cdot \sigma_{\max}\left(1 - \frac{t}{T}\right)$\;

            Compute $p_{b,\cdot} \leftarrow \mathrm{Softmax}(z_b / \tau(t))$\;

            \tcp{Top-k head selection}
            Compute $k_b \leftarrow \max(1, \lfloor s_b H \rfloor)$\;
            Select $\mathcal{S}_b \leftarrow \mathrm{TopK}(p_{b,\cdot}, k_b)$\;

            Initialize $Y_b \leftarrow 0$\;

            \For{$h = 1$ \KwTo $H$}{

                \If{$h \in \mathcal{S}_b$}{

                    Compute attention:
                    $A_{b,h} \leftarrow \mathrm{Softmax}\left(\frac{Q_{b,h}K_{b,h}^\top}{\sqrt{d_h}}\right)V_{b,h}$\;

                    Compute weight:
                    $w_{b,h} \leftarrow s_b \cdot H \cdot p_{b,h}$\;

                    Accumulate output:
                    $Y_b \leftarrow Y_b \,\Vert\, (w_{b,h} A_{b,h})$\;

                }
                \Else{
                    $Y_b \leftarrow Y_b \,\Vert\, 0$
                }

            }

            Compute final output:
            $Y_b \leftarrow Y_b W_O$\;

            Compute task loss:
            $\mathcal{L}_{task} \leftarrow \ell(Y_b, y_b)$\;

            Compute budget loss:
            $v(s_b) \leftarrow \max(0, s_{\min} - s_b) + \max(0, s_b - s_{\max})$\;
            $\mathcal{L}_{budget} \leftarrow \alpha(s_b) \cdot v(s_b)^2$\;

            Compute entropy loss:
            $\mathcal{L}_{entropy} \leftarrow \sum_{h=1}^{H} p_{b,h} \log p_{b,h}$\;
            $\beta(t) \leftarrow \beta_{\max}\left(\frac{2t}{T} - 1\right)$\;

            \tcp{Total loss}
            $\mathcal{L} \leftarrow \mathcal{L}_{task} + \mathcal{L}_{budget} + \beta(t)\mathcal{L}_{entropy}$\;

        }

        \tcp{Parameter update}
        $\theta \leftarrow \theta - \eta \nabla_{\theta} \mathcal{L}$\;
        $\phi \leftarrow \phi - \eta \nabla_{\phi} \mathcal{L}$\;
        $W_Q, W_K, W_V, W_O \leftarrow W - \eta \nabla_{W} \mathcal{L}$\;

    }
}

\Return all parameters;
\end{algorithm}

\noindent The model is fully differentiable during training, while sparsity is applied only at inference time. The overall procedure of BudgetFormer is summarized in Algorithm~\ref{alg:budgetformer}.

\subsection{Complexity Analysis}

We analyze the computational and memory complexity of BudgetFormer and compare it to standard multi-head attention.\newline
In standard attention, the cost of a single layer is dominated by:
\begin{equation}
\mathcal{C}_{\text{MHA}}(X) \approx \mathcal{O}(B N D^2) + \mathcal{O}(B H N^2 d_h),
\end{equation}
where all $H$ heads are computed for every input. The second term dominates and scales linearly with $H$.\newline

\noindent In BudgetFormer, additional computations arise from the budget and gating networks:
\begin{equation}
\mathcal{C}_{\text{budget}} \approx \mathcal{O}(B D^2) + \mathcal{O}(B D H),
\end{equation}
which are independent of the sequence length $N$ and negligible compared to the attention cost.\newline
During training, all heads are evaluated:
\begin{equation}
\mathcal{C}_{\text{train}} \approx \mathcal{C}_{\text{MHA}} + \mathcal{C}_{\text{budget}},
\end{equation}
ensuring stable gradients and full exploration of the head space. The overhead induced by $f_{\theta}$ and $g_{\phi}$ remains marginal relative to the quadratic attention term.\newline

\noindent During inference, only the top-$k$ heads are computed, with:
\begin{equation}
k = \lfloor s \cdot H \rfloor.
\end{equation}
The attention cost becomes:
\begin{equation}
\mathcal{C}_{\text{inference}} \approx \mathcal{O}(B N D^2) + \mathcal{O}(B k N^2 d_h).
\end{equation}
This yields a proportional reduction:
\begin{equation}
\frac{\mathcal{C}_{\text{inference}}}{\mathcal{C}_{\text{MHA}}} \approx \frac{k}{H} = s.
\end{equation}
Hence, the computational cost scales linearly with the predicted budget $s$, allowing input-dependent efficiency.\newline

\noindent A similar reduction applies to memory. In standard attention:
\begin{equation}
\mathcal{M}_{\text{MHA}}(X) \approx \mathcal{O}(B H N^2),
\end{equation}
while BudgetFormer reduces this to:
\begin{equation}
\mathcal{M}_{\text{inference}}(X) \approx \mathcal{O}(B k N^2),
\end{equation}
leading to:
\begin{equation}
\frac{\mathcal{M}_{\text{inference}}}{\mathcal{M}_{\text{MHA}}} \approx s.
\end{equation}
This reduction directly translates into lower memory usage and improved scalability for long sequences.\newline

\noindent Finally, since energy consumption is approximately proportional to the number of floating-point operations, BudgetFormer also reduces the inference-time carbon footprint:
\begin{equation}
\text{CO}_2 \propto \mathcal{C}_{\text{inference}} \propto s.
\end{equation}
Overall, BudgetFormer preserves the full expressivity of multi-head attention during training while enabling a controllable and input-adaptive reduction in computation, memory, and energy usage at inference time.

\section{Experiments}

\subsection{Experimental Setup}
We evaluate BudgetFormer on text classification tasks by comparing it to a standard Transformer encoder using full multi-head attention. All models are trained and evaluated on five widely used benchmark datasets, covering a range of domains and classification granularities.\newline

\noindent The datasets used in our experiments are summarized in Table~\ref{tab:datasets}. They include topic classification (DBpedia, AG News), sentiment analysis (IMDB, Yelp Review Full), and natural language inference (SNLI). For datasets without an official validation split, we use the test set as validation.\newline

\begin{table*}[h]
\centering
\begin{tabular}{l l c c c c}
\hline
Dataset & Description & Train & Val & Test & Classes \\
\hline
DBpedia~\cite{zhang2015character} & Ontology classification & 560,000 & 70,000 & 70,000 & 14 \\
AG News~\cite{zhang2015character} & News topic classification & 120,000 & 7,600 & 7,600 & 4 \\
IMDB~\cite{maas2011learning} & Sentiment analysis & 25,000 & 25,000 & 25,000 & 2 \\
SNLI~\cite{bowman2015large} & Natural language inference & 549,367 & 9,842 & 9,824 & 3 \\
Yelp Full~\cite{zhang2015character} & Review rating prediction & 650,000 & 50,000 & 50,000 & 5 \\
\hline
\end{tabular}
\caption{Datasets used for evaluation.}
\label{tab:datasets}
\end{table*}
\noindent As a baseline, we use a Transformer encoder composed of $L=4$ layers, each with $H=8$ attention heads and model dimension $D=768$. BudgetFormer follows the same architecture, replacing the standard attention layer with the proposed adaptive head budgeted attention.\newline
The budget predictor $f_{\theta}$ is implemented as a two-layer feed-forward network with a ReLU activation, mapping $\mathbb{R}^{D} \rightarrow \mathbb{R}$. The head scoring function $g_{\phi}$ is a single linear projection mapping $\mathbb{R}^{D} \rightarrow \mathbb{R}^{H}$.\newline
Both models are trained for 10 epochs using the AdamW optimizer with a learning rate of $2 \times 10^{-5}$ and a batch size of 16. The number of training steps $T$ depends on the dataset size and is used consistently in the scheduling functions defined in Section~\ref{section:method}.\newline

\noindent For BudgetFormer, the budget is constrained within $[s_{\min}, s_{\max}] = [0.1, 0.9]$, allowing the model to explore a wide range of computational allocations without bias toward extreme values.\newline
The training hyperparameters are set as follows: $\alpha_{\text{base}} = 0.001$, $\alpha_{\max} = 0.05$, $\beta_{\max} = 0.05$, $\sigma_{\max} = 0.5$, $\tau_{\max} = 2.0$, $\tau_{\min} = 0.1$, and $\gamma = 5.0$.\newline

\noindent In terms of model size, the baseline Transformer requires 197.58 MB of memory, while BudgetFormer requires 206.70 MB. This increase is due to the additional parameters introduced by $f_{\theta}$ and $g_{\phi}$, which remain lightweight compared to the attention layers and introduce negligible computational overhead.\newline
All experiments are conducted on a single NVIDIA A100 GPU with 80GB of memory.

\subsection{Main Results}

We report the main results on five text classification benchmarks in Table~\ref{tab:main_results}. We compare BudgetFormer against a standard Transformer encoder with identical architecture and training setup. For BudgetFormer, inference is performed using top-$k$ head selection, where $k = \lfloor s \cdot H \rfloor$, enabling actual computational savings.\newline
\begin{table*}[t]
\centering
\small
\begin{tabular}{l|cc|cc|cc|c}
\hline
\textbf{Dataset} & \multicolumn{2}{c|}{\textbf{Accuracy}} & \multicolumn{2}{c|}{\textbf{FLOPs (Test)}} & \multicolumn{2}{c|}{\textbf{Carbon (gCO2)}} & \textbf{$s_{\text{mean}}$} \\
 & Transformer & BudgetFormer & Transformer & BudgetFormer & Transformer & BudgetFormer &  \\
\hline
DBpedia & 0.9830 & \textbf{0.9859} & $4.40\cdot10^{13}$ & $\mathbf{4.26\cdot10^{13}}$ & 0.1468 & \textbf{0.1420} & 0.085 \\
AG News & \textbf{0.9099} & 0.9022 & $4.78\cdot10^{12}$ & $\mathbf{4.65\cdot10^{12}}$ & 0.0159 & \textbf{0.0155} & 0.212 \\
IMDB & 0.8354 & \textbf{0.8356} & $1.61\cdot10^{14}$ & $\mathbf{1.45\cdot10^{14}}$ & 0.5376 & \textbf{0.4840} & 0.601 \\
SNLI & 0.7835 & \textbf{0.8106} & $6.18\cdot10^{12}$ & $\mathbf{6.05\cdot10^{12}}$ & 0.0206 & \textbf{0.0202} & 0.364 \\
Yelp & 0.5810 & \textbf{0.6190} & $3.23\cdot10^{14}$ & $\mathbf{2.58\cdot10^{14}}$ & 1.0751 & \textbf{0.8591} & 0.198 \\
\hline
\end{tabular}
\caption{Comparison between standard Transformer and BudgetFormer on test sets. BudgetFormer uses adaptive head selection at inference (top-$k$). FLOPs and carbon correspond to full evaluation over the test set.}
\label{tab:main_results}
\end{table*}

\noindent BudgetFormer consistently achieves competitive or improved performance compared to the standard Transformer, while reducing inference cost. On DBpedia and SNLI, we observe clear accuracy gains, with improvements of +0.29 and +2.71 points respectively. On Yelp, the gain is even more pronounced (+3.8 points), suggesting that adaptive head selection is particularly beneficial for more complex or noisy datasets.\newline

\noindent On AG News, performance remains close to the baseline with a slight drop (-0.7), while still reducing computational cost. This indicates that for simpler datasets, aggressive head reduction may slightly affect performance, but remains controlled.\newline

\noindent From an efficiency perspective, BudgetFormer systematically reduces FLOPs and carbon emissions at inference. The reduction is directly correlated with the learned budget $s_{\text{mean}}$. For instance, on DBpedia, the model uses only 8.5\% of heads on average, leading to lower computational cost with improved accuracy. On IMDB, where longer sequences require richer representations, the model allocates a higher budget ($s \approx 0.6$), preserving performance while still reducing cost by approximately 10\%.\newline
Importantly, these gains are achieved without modifying the training pipeline. During training, all heads remain active, ensuring stable optimization and full gradient flow. The additional overhead introduced by the budget and gating networks is negligible compared to the overall model size (approximately +9 MB in parameters), and does not significantly impact training cost.\newline

\noindent Overall, these results demonstrate that BudgetFormer effectively adapts computational resources to input complexity, achieving a favorable trade-off between accuracy and efficiency. The model learns when fewer heads are sufficient and when more capacity is required, leading to both improved generalization and reduced inference cost.

\subsection{Efficiency Analysis}
We analyze the behavior of BudgetFormer along two complementary dimensions: (i) the evolution of the learned budget during training, and (ii) its adaptation to input complexity at inference for text classification (DBpedia), sentiment analysis (Yelp), and natural language inference (SNLI).\newline

\begin{figure*}[t]
\centering

\begin{subfigure}{0.32\textwidth}
    \centering
    \includegraphics[width=\linewidth]{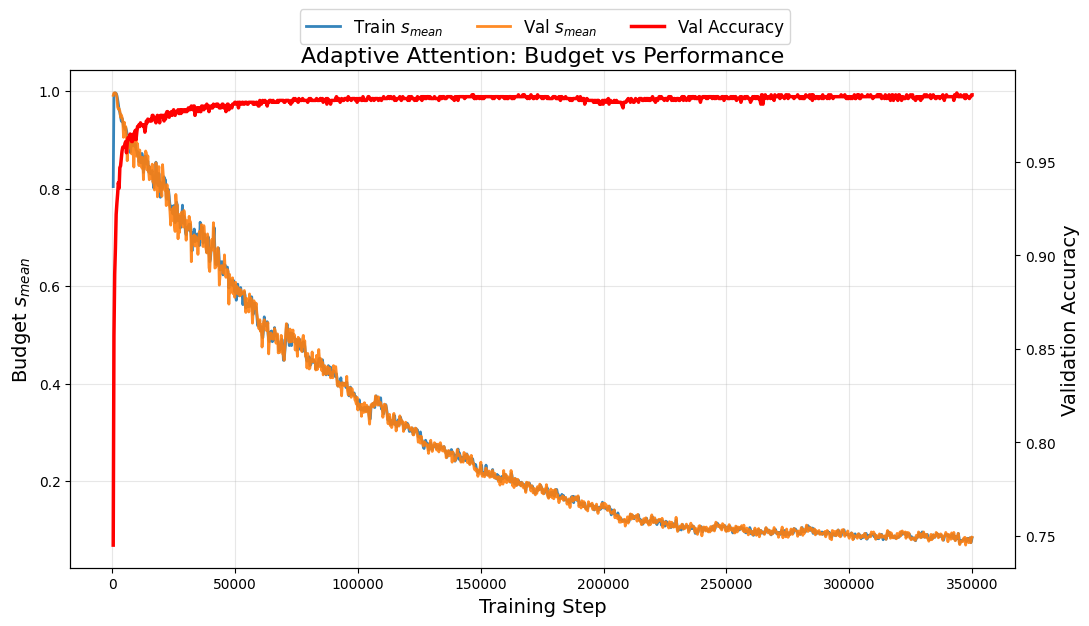}
    \caption{DBpedia}
\end{subfigure}
\hfill
\begin{subfigure}{0.32\textwidth}
    \centering
    \includegraphics[width=\linewidth]{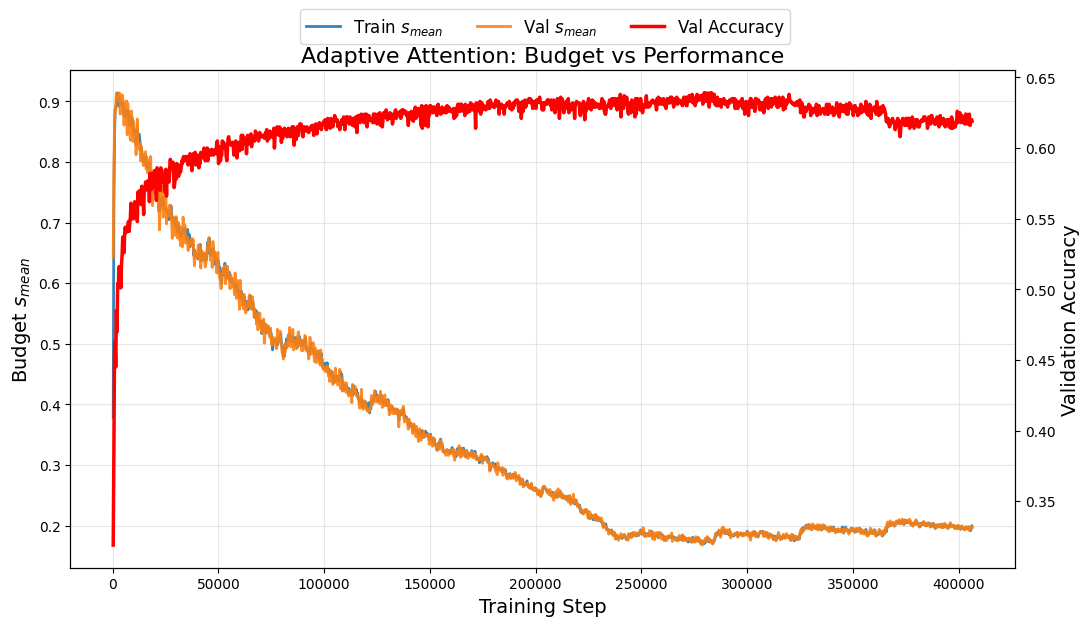}
    \caption{Yelp}
\end{subfigure}
\hfill
\begin{subfigure}{0.32\textwidth}
    \centering
    \includegraphics[width=\linewidth]{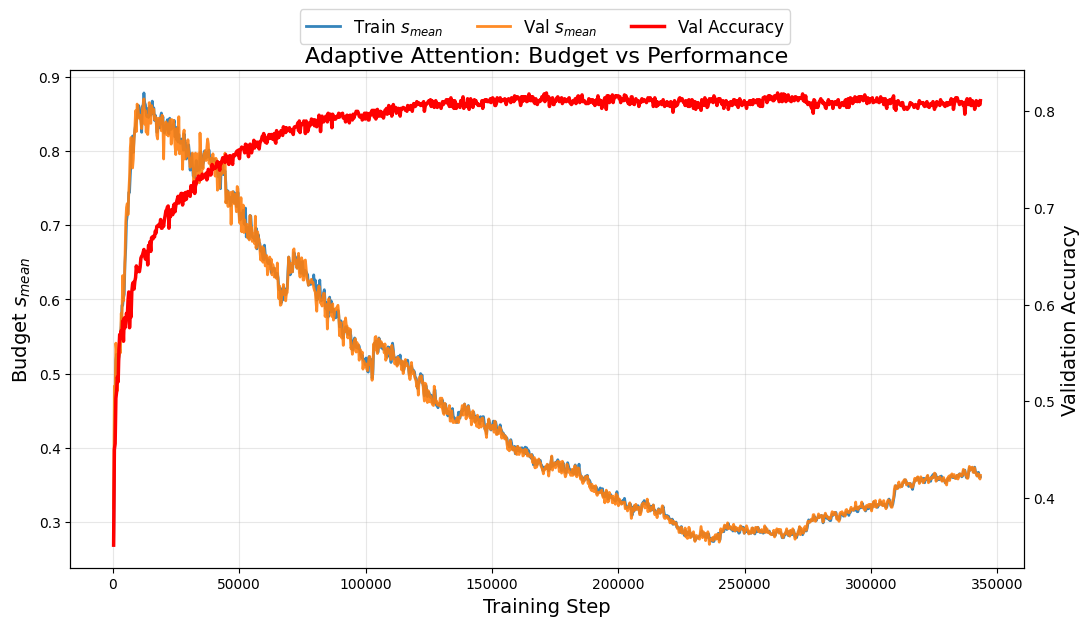}
    \caption{SNLI}
\end{subfigure}

\caption{
Training dynamics showing the evolution of $s_{\text{mean}}$ (train and validation) and validation accuracy over epochs for text classification (DBpedia), sentiment analysis (Yelp), and natural language inference (SNLI).
}
\label{fig:training_dynamics}
\end{figure*}

\noindent \textbf{Training dynamics.}
We first study the evolution of the average budget $s_{\text{mean}}$ on both training and validation sets, jointly with the validation accuracy. Figure~\ref{fig:training_dynamics} reports these curves for representative datasets.
At early stages of training, the budget $s_{\text{mean}}$ is relatively high, reflecting an exploration phase where multiple attention heads are actively used. As training progresses, $s_{\text{mean}}$ gradually decreases, indicating a transition toward a more selective and efficient allocation of heads.\newline
Importantly, we observe a strong alignment between training and validation curves, with no noticeable gap. This suggests that the learned budget generalizes well and does not overfit to the training data.\newline
At the same time, the validation accuracy steadily improves and remains stable as $s_{\text{mean}}$ decreases. This indicates that reducing the number of active heads does not harm performance. On the contrary, the model learns to discard redundant heads while preserving or improving predictive accuracy, highlighting an effective transition from exploration to exploitation.\newline

\begin{figure*}[t]
\centering

\begin{subfigure}{0.31\textwidth}
    \centering
    \includegraphics[width=\linewidth]{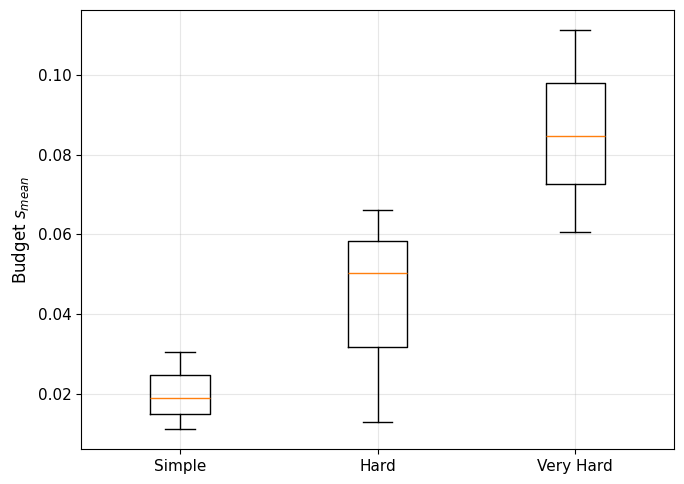}
    \caption{DBpedia}
\end{subfigure}
\hfill
\begin{subfigure}{0.31\textwidth}
    \centering
    \includegraphics[width=\linewidth]{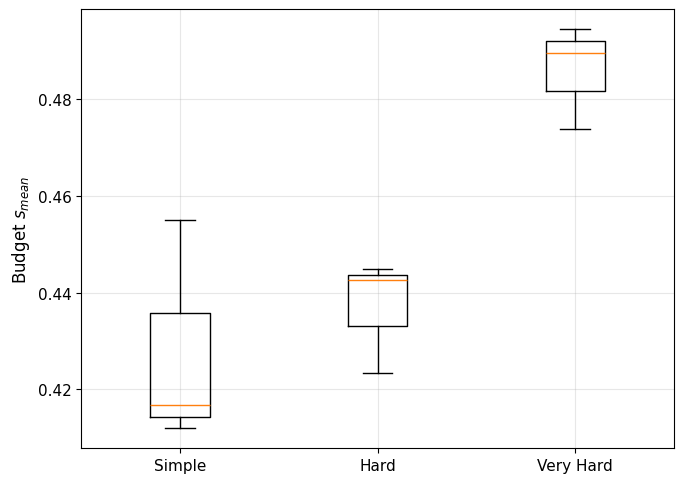}
    \caption{Yelp}
\end{subfigure}
\hfill
\begin{subfigure}{0.31\textwidth}
    \centering
    \includegraphics[width=\linewidth]{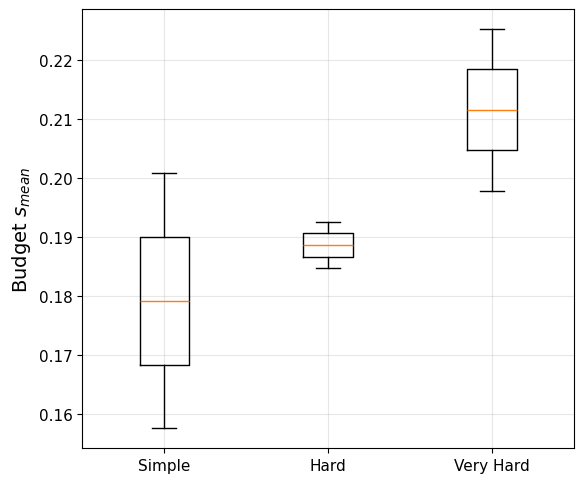}
    \caption{SNLI}
\end{subfigure}

\caption{
Distribution of the predicted budget $s$ across input complexity levels (Simple, Hard, and Very Hard) for text classification (DBpedia), sentiment analysis (Yelp), and natural language inference (SNLI).
}
\label{fig:budget_boxplots}
\end{figure*}

\noindent \textbf{Adaptation to input complexity.}
We then evaluate how the predicted budget varies with input difficulty.
For each dataset, we construct three categories of inputs: \textit{Simple}, \textit{Hard}, and \textit{Very hard}. Figure~\ref{fig:budget_boxplots} presents the distribution of $s$ across these categories.

\noindent Across all datasets, we observe a consistent increase of $s$ with input complexity. Simple inputs require only a small fraction of attention heads, while more complex inputs trigger higher budgets.\newline
This trend is particularly clear on SNLI, where logically challenging examples require more heads, and on Yelp, where nuanced sentiment leads to higher computational demand.\newline
These results demonstrate that BudgetFormer effectively adapts its computational effort to the input. The model allocates more resources when necessary while remaining efficient on simpler examples, leading to a form of conditional computation at the head level.

\subsection{Ablation Study}

We conduct two complementary ablations to isolate the roles of the budget predictor $f_{\theta}$ and the head selection network $g_{\phi}$.\newline

\noindent \textbf{Ablation 1: Fixed budget (no $f_{\theta}$).}  
We remove the learned budget and fix $s \in \{0.1, 0.25, 0.5, 0.75, 1.0\}$ while keeping $g_{\phi}$ trainable. Results are reported in Table~\ref{tab:ablation_fixed_budget}.  
\begin{table}[h]
\centering
\small
\begin{tabular}{l|c|c|c|c|c}
\hline
\textbf{Dataset} & $s=0.1$ & $s=0.25$ & $s=0.5$ & $s=0.75$ & $s=1.0$ \\
\hline
DBpedia & 0.9846 & 0.9478 & 0.6202 & 0.3245 & 0.1670  \\
AG News & 0.8905 & 0.8974 & 0.8675 & 0.8139 & 0.7554 \\
IMDB & 0.7296 & 0.8036 & 0.8246 & 0.8289 & 0.8162 \\
SNLI & 0.6704 & 0.7598 & 0.5838 & 0.3541 & 0.3190 \\
Yelp & 0.5818 & 0.5813 & 0.2737 & 0.1013 & 0.0743 \\
\hline
\end{tabular}
\caption{Accuracy with fixed budget $s$ (no learned $f_{\theta}$).}
\label{tab:ablation_fixed_budget}
\end{table}
We observe a strong degradation when $s$ increases on several datasets (DBpedia, SNLI, Yelp). This shows that allocating more heads does not necessarily improve performance. Without adaptive control, larger budgets introduce noise through $g_{\phi}$, leading to inefficient head utilization. In contrast, BudgetFormer (Table~\ref{tab:main_results}) learns small but optimal budgets (e.g., $s_{\text{mean}} \approx 0.085$ on DBpedia), achieving higher accuracy with fewer active heads.\newline

\noindent \textbf{Ablation 2: Random head selection (no learned $g_{\phi}$).}  
We fix $s$ to the learned value from BudgetFormer and replace $g_{\phi}$ with random head selection. Results are shown in Table~\ref{tab:ablation_random_gate}.
\begin{table}[h]
\centering
\small
\begin{tabular}{l|c|c}
\hline
\textbf{Dataset} & Random $g_{\phi}$ & BudgetFormer \\
\hline
DBpedia & 0.7511 & \textbf{0.9859} \\
AG News & 0.8180 & \textbf{0.9022} \\
IMDB & 0.7493 & \textbf{0.8356} \\
SNLI & 0.3370 & \textbf{0.8106} \\
Yelp & 0.3244 & \textbf{0.6190} \\
\hline
\end{tabular}
\caption{Impact of removing learned head selection $g_{\phi}$ (random gating).}
\label{tab:ablation_random_gate}
\end{table}
The performance collapses across all datasets when head selection is random, even with the correct budget. This demonstrates that $g_{\phi}$ is essential to identify relevant heads. The budget $s$ alone is insufficient: performance depends on \emph{which} heads are selected, not only how many.\newline

\noindent These ablations highlight two key properties:  
(i) the budget must be \emph{adaptive} (learned via $f_{\theta}$), as fixed allocations are suboptimal and can introduce noise,  
(ii) head selection must be \emph{structured} (learned via $g_{\phi}$), as random selection severely degrades performance.\newline

Together, $f_{\theta}$ and $g_{\phi}$ enable BudgetFormer to allocate computation both \emph{quantitatively} (how many heads) and \emph{qualitatively} (which heads), explaining the gains observed in Table~\ref{tab:main_results}.

\subsection{Generalization Across Model and Data Scales}

We evaluate whether BudgetFormer maintains its advantages when scaling (i) the model capacity and (ii) the amount of training data. All experiments are conducted on SNLI for controlled comparison.\newline

\noindent \textbf{Scaling model capacity.}  
We vary both the number of layers and attention heads, and compare against a standard Transformer of identical architecture. Results are summarized in Table~\ref{tab:scaling_model}.
\begin{table}[h]
\centering
\small
\begin{tabular}{l|cc|c}
\hline
\textbf{Model} & Transformer & BudgetFormer & $s_{\text{mean}}$ \\
\hline
4L-8H  & 0.7835 & \textbf{0.8106} & 0.364 \\
4L-12H & 0.7847 & \textbf{0.8080} & 0.310 \\
6L-12H & 0.7922 & \textbf{0.8065} & 0.260 \\
12L-12H & 0.7808 & \textbf{0.8193} & 0.105 \\
\hline
\end{tabular}
\caption{Scaling model depth (L) and heads (H) on SNLI.}
\label{tab:scaling_model}
\end{table}
BudgetFormer consistently outperforms the Transformer across all configurations. A key observation is that $s_{\text{mean}}$ decreases as model capacity increases. For instance, with 12 layers and 12 heads, the model achieves its best accuracy (0.8193) while using only $\sim$10\% of the heads on average. This indicates that larger models contain redundant heads, and BudgetFormer effectively exploits this redundancy by selecting only the most relevant ones. In contrast, the standard Transformer does not benefit as much from scaling, suggesting inefficient use of additional capacity.\newline

\noindent \textbf{Scaling data size.}  
We now vary the fraction of the training set used ($10\%$, $25\%$, $50\%$, $100\%$). Results are shown in Table~\ref{tab:scaling_data}.
\begin{table}[h]
\centering
\small
\begin{tabular}{l|cc|c}
\hline
\textbf{Data fraction} & Transformer & BudgetFormer & $s_{\text{mean}}$ \\
\hline
10\% & 0.6151 & \textbf{0.6158} & 0.445 \\
25\% & 0.6814 & \textbf{0.7075} & 0.427 \\
50\% & 0.7243 & \textbf{0.7738} & 0.388 \\
100\% & 0.7835 & \textbf{0.8106} & 0.364 \\
\hline
\end{tabular}
\caption{Scaling training data size on SNLI.}
\label{tab:scaling_data}
\end{table}
BudgetFormer shows stronger robustness in low-data regimes. At $10\%$ of the data, both models perform similarly, but BudgetFormer quickly surpasses the Transformer as more data becomes available. Notably, $s_{\text{mean}}$ adapts to the data regime: it is higher when data is scarce (0.445 at 10\%), indicating that the model uses more heads to compensate for uncertainty, and decreases as more data becomes available (down to 0.388 at 50\%). This reflects an adaptive trade-off between exploration and efficient computation.\newline

\noindent These results highlight two important properties. First, BudgetFormer scales better with model capacity by avoiding redundant computation and focusing on a subset of useful heads. Second, it adapts its computational budget to the amount of available data, using more resources when necessary and becoming more selective as learning stabilizes. This dynamic behavior leads to consistently better accuracy while maintaining controlled computational cost, demonstrating strong generalization across both model and data scales.

\subsection{Qualitative Analysis}
We conduct a qualitative analysis on SNLI to better understand the behavior of the scaling factor $s$ and the head selection distribution $q$ across transformer blocks and inference labels. SNLI is particularly interesting because natural language inference requires modeling semantic relationships between a premise and a hypothesis, often involving more complex reasoning than single-sentence classification tasks. This makes it a suitable benchmark for analyzing how BudgetFormer allocates computational resources across layers and inputs.\newline

\begin{figure*}[t]
\centering

\begin{subfigure}{0.32\textwidth}
    \centering
    \includegraphics[width=\linewidth]{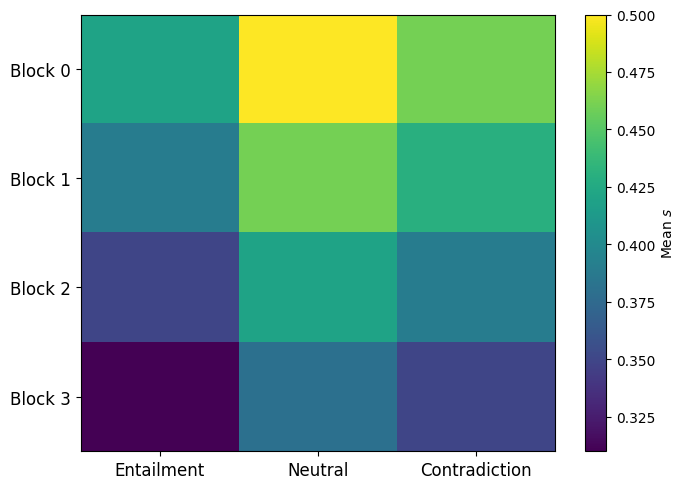}
    \caption{Mean budget $s$.}
\end{subfigure}
\hfill
\begin{subfigure}{0.32\textwidth}
    \centering
    \includegraphics[width=\linewidth]{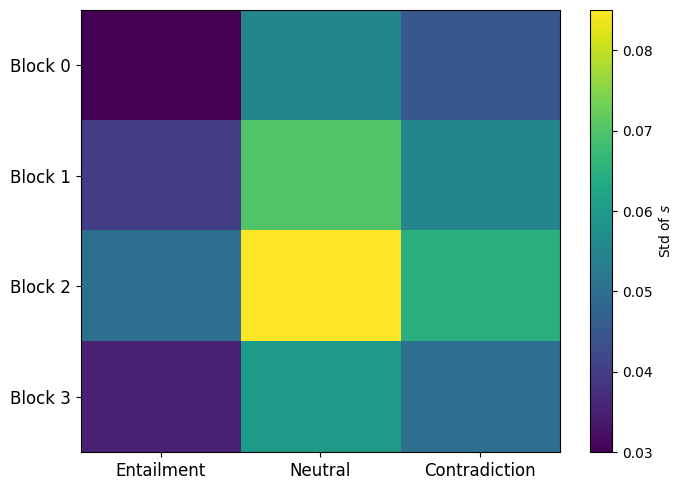}
    \caption{Standard deviation of $s$.}
\end{subfigure}
\hfill
\begin{subfigure}{0.32\textwidth}
    \centering
    \includegraphics[width=\linewidth]{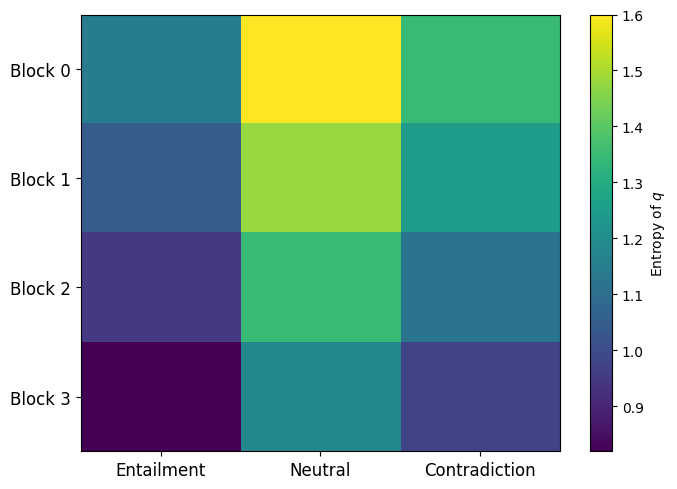}
    \caption{Entropy of $q$.}
\end{subfigure}

\caption{
Analysis of BudgetFormer on SNLI across transformer blocks and inference labels. (a) Mean predicted budget $s$, (b) standard deviation of $s$, and (c) entropy of the head-selection distribution $q$. Higher budgets and entropy values are observed for neutral examples, indicating increased computational requirements and more distributed head utilization compared to entailment and contradiction examples.
}
\label{fig:snli_heatmaps}
\end{figure*}

\noindent \textbf{Budget allocation, variability, and head-selection entropy.}
Figure~\ref{fig:snli_heatmaps} summarizes three complementary aspects of the adaptive allocation mechanism across transformer blocks and inference labels. Figure~\ref{fig:snli_heatmaps}(a) reports the mean predicted budget $s$, Figure~\ref{fig:snli_heatmaps}(b) shows the standard deviation of $s$, and Figure~\ref{fig:snli_heatmaps}(c) presents the entropy of the head-selection distribution $q$.
A first observation is that the predicted budget varies consistently across inference labels. Neutral examples exhibit the highest $s_{mean}$ values across all blocks, while entailment examples require the lowest budget. Contradiction occupies an intermediate regime. This behavior is consistent with the nature of the task: neutral pairs often involve semantic ambiguity and multiple plausible interpretations, whereas entailment examples can frequently be resolved from more direct semantic evidence.\newline

\noindent The standard deviation of $s$ in Figure~\ref{fig:snli_heatmaps}(b) further shows that budget allocation is input-dependent rather than fixed. Variability is particularly pronounced for neutral examples and reaches its maximum in intermediate layers, suggesting that the model dynamically adjusts its computational allocation when semantic relationships are less certain.\newline

\noindent The entropy analysis in Figure~\ref{fig:snli_heatmaps}(c) reveals a similar pattern. Neutral examples exhibit higher entropy values, indicating that evidence is distributed across a larger number of attention heads. In contrast, entailment examples show lower entropy, suggesting that only a small subset of specialized heads is required. Entropy also decreases progressively in deeper layers, indicating increasing head specialization as representations become more refined.\newline

\noindent Taken together, these observations suggest that BudgetFormer allocates both larger budgets and more distributed head usage to semantically ambiguous inputs, while relying on fewer specialized heads for simpler inference cases. This behavior supports the hypothesis that the proposed adaptive mechanism effectively adjusts computational resources to the complexity of the input.\newline

\begin{figure*}
    \centering
    \includegraphics[width=\linewidth]{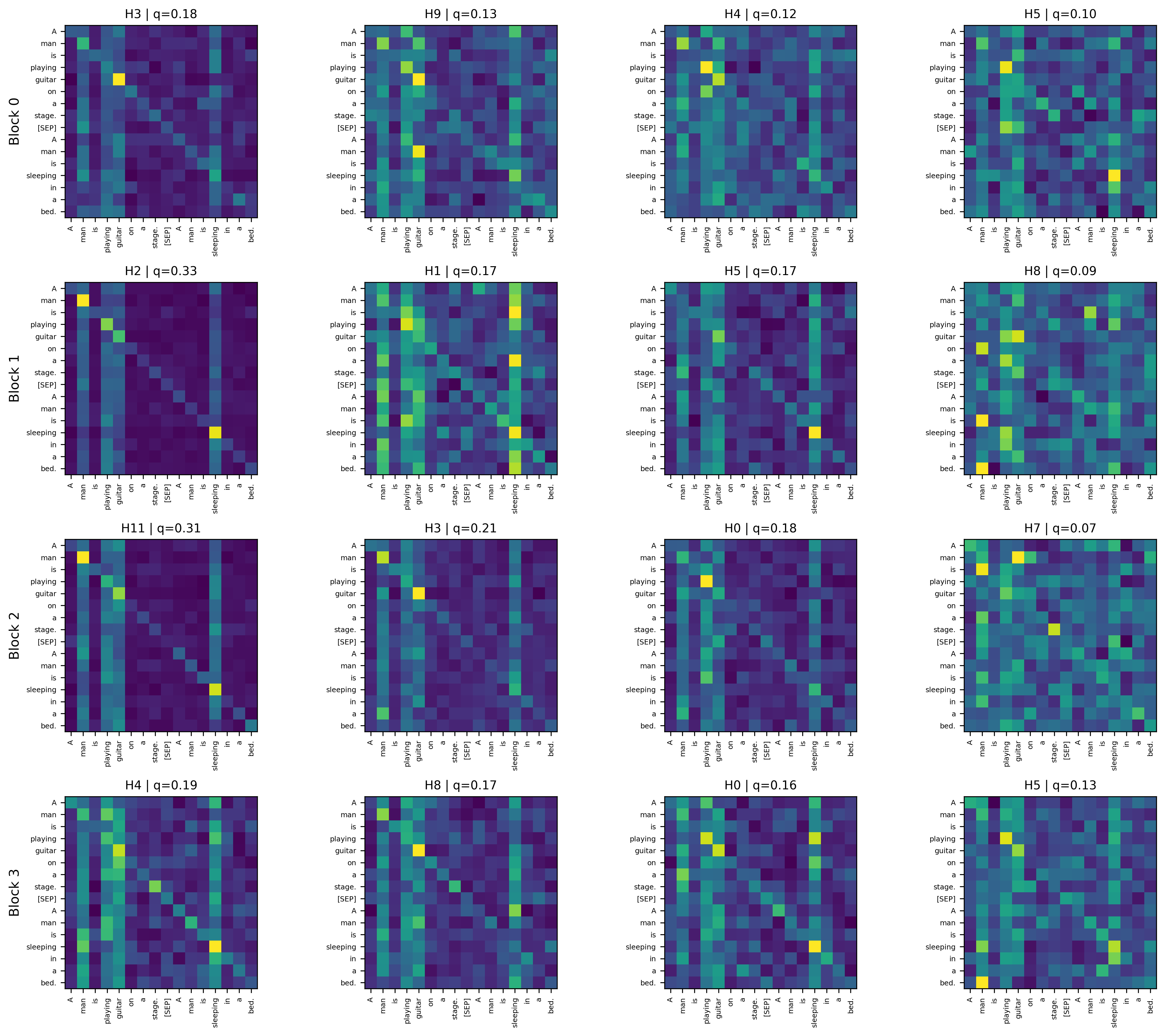}
    \caption{Attention maps across transformer blocks for a representative SNLI example (premise: "A man is playing guitar on a stage."; hypothesis: "A man is sleeping in a bed."). For each block, only the top-4 heads ranked by importance score $q$ are visualized, consistent with the average budget level $s_{\text{mean}} \approx 0.36$. Heads are sorted in decreasing order of $q$, illustrating the progressive concentration of attention into a small subset of highly specialized heads across layers.}
    \label{fig:snli_attention}
\end{figure*}

\noindent \textbf{Attention visualization.}
Figure~\ref{fig:snli_attention} presents attention patterns for a representative SNLI instance: \textit{"A man is playing guitar on a stage."} (premise) and \textit{"A man is sleeping in a bed."} (hypothesis). For each transformer block, only the top-4 heads ranked by their importance score $q$ are visualized, consistent with the average budget level observed on SNLI ($s_{\text{mean}} \approx 0.36$). This reflects the fact that, under this operating regime, the model concentrates most of its computational resources on a small subset of informative heads, while the remaining heads contribute weak or noisy signals.\newline

\noindent Across blocks, a clear specialization pattern emerges: heads with higher $q$ values produce sharper and more semantically aligned attention maps, capturing key token interactions between premise and hypothesis (notably \textit{man}, \textit{playing}, and \textit{sleeping}). In contrast, lower-ranked heads exhibit more diffuse and less structured attention, indicating limited contribution to the decision process. This effect becomes increasingly pronounced in deeper blocks, where attention is progressively concentrated into a reduced number of highly specialized heads.\newline

\noindent Overall, these observations support the effectiveness of the adaptive head budgeting mechanism: it induces structured sparsity in head utilization, where a small set of dominant heads carries most of the semantic information required for inference, while redundant heads are effectively de-emphasized without degrading performance.

\section{Limitations}

While the proposed approach shows strong performance on coarse-grained classification tasks, it has several limitations.
A first limitation stems from the use of global pooling in $f_\theta$ and $g_\phi$, which aggregates token representations uniformly. This design assumes equal contribution from all tokens when estimating the budget $s$ and head distribution $q$, and therefore does not explicitly model token-level heterogeneity.
This simplification is appropriate for tasks where global semantics dominate, but becomes restrictive for more complex settings such as question answering or fine-grained reasoning, where only a subset of tokens carries task-relevant information.
A second limitation is that the framework operates exclusively at the head level. While this enables structured sparsity in attention, it ignores potential redundancy at the token level, which could further improve computational efficiency.\newline

\noindent Overall, these constraints limit expressiveness in scenarios requiring localized or structured reasoning, and suggest that future work should explore token-aware budgeting mechanisms that remain efficient while capturing finer-grained importance patterns.

\section{Conclusion}
This work introduced a dynamic scaling mechanism for Transformer attention, enabling input-dependent modulation of attention heads through learned budget and head importance distributions. The model consistently learns to concentrate computation on a small subset of heads, yielding sparse and efficient representations.
Through qualitative analysis on SNLI, several key properties were observed: (i) the budget variable $s$ adapts across layers and inputs, (ii) head importance is highly concentrated on a few dominant heads, and (iii) deeper layers exhibit stronger sparsification, consistent with reduced redundancy in higher-level representations. Overall, these results indicate that sparsity naturally emerges in multi-head attention and can be effectively leveraged through adaptive computation.\newline
\noindent \textbf{Future Work.}
Future directions include extending the proposed mechanism to more complex tasks such as long-context reasoning and question answering, where fine-grained token interactions are critical. Another promising direction is integrating adaptive head budgeting into large-scale language models to evaluate its scalability in modern architectures. Finally, combining head-level sparsity with token-level pruning could further reduce computational cost while preserving performance, enabling more efficient Transformer variants.

\bibliographystyle{ieeetr}
\bibliography{bibliography}

\clearpage

\end{document}